\documentclass{article}

\usepackage{arxiv}

\usepackage{amsmath,amsfonts,amssymb}
\usepackage{array}
\usepackage{booktabs}
\usepackage{pdfpages}
\usepackage{multirow}
\usepackage{tabularx}
\usepackage{graphicx}
\usepackage[caption=false,font=normalsize,labelfont=sf,textfont=sf]{subfig}
\usepackage{textcomp}
\usepackage{stfloats}
\usepackage{url}
\usepackage{verbatim}
\usepackage{cite}
\usepackage{algorithm}
\usepackage{orcidlink}
\usepackage{algpseudocode}
\usepackage{pgfplots}
\usepackage{graphicx}
\usepackage{wrapfig}
\usepackage{pgfplotstable}
\usepackage{graphicx}
\usepackage{multirow}
\usepackage{amsmath,amssymb,amsfonts}
\usepackage{amsthm}
\usepackage{mathrsfs}
\usepackage[title]{appendix}
\usepackage{xcolor}
\usepackage{textcomp}
\usepackage{manyfoot}
\usepackage{booktabs}
\usepackage{algorithm}
\usepackage{algorithmicx}
\usepackage{algpseudocode}
\usepackage{listings}
\usepackage[utf8]{inputenc}
\usepackage[T1]{fontenc}
\usepackage{hyperref}
\usepackage{orcidlink}
\usepackage{array}
\usepackage{tabularx}
\usepackage{url}
\usepackage{enumitem}
\usepackage{pgfplots}
\usepackage{tikz}

\pgfplotsset{compat=1.18}
\usetikzlibrary{arrows.meta,positioning,calc}

\definecolor{designBlue}{HTML}{2563EB}
\definecolor{methodPurple}{HTML}{7C3AED}
\definecolor{conditionOrange}{HTML}{EA580C}
\definecolor{evidenceGreen}{HTML}{15803D}
\definecolor{provTeal}{HTML}{0F766E}
\definecolor{auditRed}{HTML}{B91C1C}
\definecolor{analysisIndigo}{HTML}{4338CA}
\definecolor{arrowDark}{HTML}{374151}
\definecolor{textDark}{HTML}{111827}

\title{\textbf{Analysis of Federated Aggregation under Model Poisoning and Backdoor Attacks}: A Reconstructed Cross-Dataset and Cross-Architecture Benchmark}

\author{
\textbf{Soumya Mazumdar}~\orcidlink{0009-0006-3521-9557}\\
Department of Computer Science and Business Systems\\
\textbf{\textit{Gargi Memorial Institute of Technology}}\\
Affiliated to Maulana Abul Kalam Azad University of Technology\\
Balarampur, Mouza Beralia, Baruipur, Kolkata 700144, West Bengal, India\\
\texttt{reachme@soumyamazumdar.com}
\and
\textbf{Vineet Kumar Rakesh}~\orcidlink{0009-0000-7102-6564}\\
Computer and Informatics Group, \textbf{\textit{Variable Energy Cyclotron Centre}}\\
1/AF, Bidhannagar, Kolkata 700064, West Bengal, India\\
Engineering Sciences, \textbf{\textit{Homi Bhabha National Institute}}\\
Anushaktinagar, Mumbai, Maharashtra 400094, India\\
\texttt{vineet@vecc.gov.in}
\and
\textbf{Tapas Samanta}~\orcidlink{0000-0003-0521-0747}\\
Computer and Informatics Group, \textbf{\textit{Variable Energy Cyclotron Centre}}\\
1/AF, Bidhannagar, Kolkata 700064, West Bengal, India\\
Engineering Sciences, \textbf{\textit{Homi Bhabha National Institute}}\\
Anushaktinagar, Mumbai, Maharashtra 400094, India\\
\texttt{tsamanta@vecc.gov.in}
}

\begin{document}
\maketitle

\begin{abstract}
Robust comparisons of federated aggregation methods require joint consideration of predictive performance, threat definitions, metric semantics, and execution provenance. A 500-cell seed-1 evaluation matrix was reconstructed across five aggregation methods, five datasets, five architectures, and four recorded conditions: clean, sign-flipping, Gaussian, and BadNets. Each cell contains a numerical performance summary. Successful execution logs were identified for 454 original runs and 36 repaired or rerun executions, whereas 10 clean SVHN cells were supported by summary-only provenance. Trimmed Mean achieved the highest clean macro-mean accuracy (76.02\%) and the lowest mean within-task rank (1.70). Krum attained the highest recorded accuracy under both sign-flipping and Gaussian configurations. These relative rankings remained unchanged when analysis was restricted to 21 task pairs for which original successful logs were available for every method–condition combination. Audit of the supplied BadNets metric implementation established that every test input is triggered prior to target-label counting; consequently, the retained metric represents Triggered Target-Label Rate (TTLR) rather than a conventional target-excluding attack success rate. An audit of the supplied FedPARETO scaffold further identified a pathway in which predictive summaries may characterise an uncorrupted local model while the aggregation weight is applied to a separately corrupted update, introducing a potential discrepancy between reported predictive outcomes and the updates used for aggregation. The canonical matrix contains a single identified seed for each cell, and the exact attack and configuration lineage is incomplete. Accordingly, the reported findings should be interpreted as descriptive comparisons within the recorded configurations and should not be construed as statistical estimates or universal claims regarding robustness.
\end{abstract}


\section{Introduction}

Federated learning (FL) learns a shared model using dispersed client data without needing the server to gather the clients' raw training samples. In the conventional FedAvg process, a global model is broadcast, clients execute local optimization, and the server merges the resultant model updates, generally with sample-size weighting~\cite{McMahan2017}. This design modifies the server's observability: the server gets parameter updates rather than the underlying local data-generating operations.  Consequently, aggregation must balance beneficial but diverse client updates against changes that are erroneous, unusual, or purposely damaging.

This problem is confounded by statistical heterogeneity. Honest customers might maximise distinct empirical aims because class proportions, acquisition circumstances, sample numbers, or domains vary between clients. FedProx, SCAFFOLD, FedNova, and adaptive federated optimization handle distinct kinds of client drift and objective inconsistency \cite{Li2020FedProx,Karimireddy2020SCAFFOLD,Wang2020FedNova,Reddi2021FedOpt}. Byzantine robustness poses a distinct problem: a malevolent participant may submit an arbitrary or strategically modified update. Sign-reversal and additive-noise attacks directly affect model updates, while backdoor attacks may retain conventional predictive performance while generating a specific reaction on trigger-bearing inputs \cite{Bagdasaryan2020}. These attack types consequently need not elicit the same observable failure signatures.

Comparisons among robust aggregation methods are difficult for three additional reasons. First, different rules rely on different information and trust assumptions. Krum uses inter-update geometry \cite{Blanchard2017}, coordinate-wise Trimmed Mean suppresses extreme coordinate values \cite{Yin2018}, and FLTrust uses a trusted server-side root update \cite{Cao2021}. Second, metric names do not guarantee metric semantics. A field labelled ``attack success rate'' can differ materially from conventional target-excluding backdoor ASR if its denominator includes naturally target-class examples. Third, historical result archives can contain unequal levels of execution provenance. Numerical summaries, successful execution logs, configuration snapshots, and source-code revisions are distinct evidence types and should not be treated as interchangeable.

\begin{figure}[ht]
    \centering
    \includegraphics[width=0.86\linewidth]{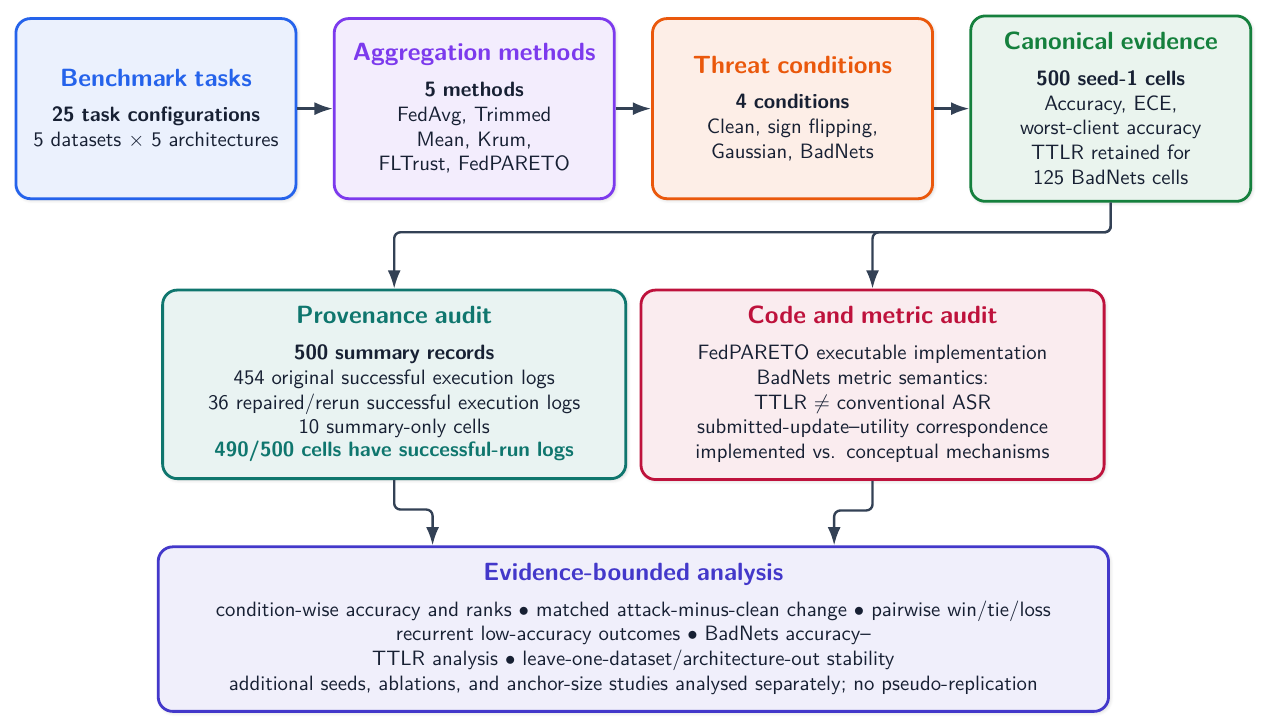}
    \caption{\textbf{Evidence reconstruction and analysis design.} The reconstructed matrix comprises 25 dataset--architecture task configurations, five aggregation methods, and four recorded conditions, yielding 500 numerical seed-1 cells. All cells have run-summary values; 454 are additionally associated with original successful execution logs, 36 with successful repaired/rerun logs, and 10 with summary-only provenance. Numerical provenance is kept separate from code and metric audit so that implementation observations are not treated as proof of the generating mechanism for every historical result.}
    \label{fig:design}
\end{figure}

The present study is therefore framed as a \emph{reconstructed comparative benchmark and evidence audit}, not as a new-algorithm superiority study. We analyze a fixed 500-cell seed-1 matrix spanning five aggregation methods, five datasets, five architectures, and four recorded condition labels. The numerical analysis is accompanied by a provenance audit, a BadNets metric-semantics audit, and an implementation audit of the supplied FedPARETO scaffold. Figure~\ref{fig:design} summarizes this separation. Importantly, source-code observations are used to characterize the audited implementation but are not assumed to establish the immutable generating revision for every historical benchmark cell.

The research covers three questions:
\begin{enumerate}[leftmargin=*]
\item \textbf{How does recorded predictive performance and within-task method ordering differ across the clean, sign-flipping, Gaussian, and BadNets condition labels, and how broadly is the observed ordering distributed across the included dataset and architecture groups?}
\item \textbf{How does ordinary BadNets test accuracy relate to the retained triggered target-label metric, and what does source-code audit establish about that metric's semantics?}
\item \textbf{What mechanisms are present in the supplied FedPARETO scaffold, and does the audited post-training poisoning pathway preserve correspondence between the predictive evidence used for utility scoring and the submitted update that receives aggregation influence?}
\end{enumerate}

The principal contributions are: (i) a complete numerical reconstruction of the 500 identified seed-1 run-summary cells and a provenance ledger distinguishing original, repaired/rerun, and summary-only evidence; (ii) descriptive reanalysis of condition-wise accuracy, within-task ranks, configuration wins, matched attack-minus-clean changes, pairwise method differences, lower-tail client performance, calibration, and leave-one-group-out sensitivity; (iii) a provenance sensitivity analysis that removes weaker-provenance task pairs without collecting new data; (iv) a source-code audit showing that the retained BadNets field has TTLR semantics under the supplied metric implementation; and (v) an implementation audit that distinguishes the executable FedPARETO heuristic from broader conceptual mechanisms and identifies an update--utility correspondence failure mode in the supplied poisoning path. These contributions concern evidence reconstruction, interpretation, and reproducibility; they do not establish universal robustness, statistical superiority, or empirical validation of an unevaluated FedPARETO refinement.

\section{Related work}

\subsection*{Federated optimization and statistical heterogeneity}
FedAvg established the iterative pattern of local optimization followed by server aggregation \cite{McMahan2017}. When local objectives differ, repeated local steps can produce client drift. FedProx constrains local deviation through a proximal term \cite{Li2020FedProx}; SCAFFOLD uses control variates \cite{Karimireddy2020SCAFFOLD}; FedNova normalizes heterogeneous local computation \cite{Wang2020FedNova}; and adaptive federated optimization introduces server-side adaptive updates \cite{Reddi2021FedOpt}. These methods illustrate why an atypical honest update need not be malicious. The reconstructed canonical benchmark spans different datasets and architectures, but it does not constitute a controlled sweep over non-IID severity; where Dirichlet $\alpha=1.0$ appears below, it is reported only for the repaired subset whose configurations explicitly retain that value.

\subsection*{Byzantine-robust aggregation}
Krum selects an update supported by neighboring updates in parameter space \cite{Blanchard2017}. Coordinate-wise robust estimators such as the median and Trimmed Mean instead suppress extreme submitted values independently for each model coordinate \cite{Yin2018}. Bulyan combines Byzantine-resilient candidate selection with coordinate-wise aggregation \cite{ElMhamdi2018}, and geometric robust aggregation has also been studied in federated settings \cite{Pillutla2022}. Bucketing methods modify the effective geometry seen by robust aggregation in heterogeneous settings \cite{Karimireddy2022Bucketing}. These approaches make different assumptions about the number and geometry of malicious and benign updates. The current benchmark contains just FedAvg, Trimmed Mean, Krum, FLTrust, and the given FedPARETO implementation; it is thus a five-method comparison rather than a full leaderboard of Byzantine defenses.

Adaptive or defense-aware poisoning is particularly critical when interpreting geometry-based approaches since an attacker might seek updates that remain plausible under the defense's selection process.  The maintained benchmark does not contain adaptive Krum-aware optimization, model replacement, or attack-strength sweeps, therefore its Krum findings are constrained to the recorded sign-flipping and Gaussian circumstances rather than extended to Byzantine assaults generally.

\subsection*{Trust and Validation aggregation}

FLTrust trains a root update from a trusted server-side dataset and utilises root alignment to weight client updates~\cite{Cao2021}. This substitutes part of peer-to-peer geometric reasoning with server-held evidence, but it includes assumptions about root-data integrity and representativeness. The given FedPARETO architecture similarly combines server-root information with predicted summaries and temporal reliability.  Such techniques need explicit accounting of which signals are autonomously calculated by the server and which are provided by clients, since client-derived validation signals may itself be manipulable under a sufficiently potent adversary.

\subsection*{Backdoor evaluation, calibration, and reproducibility}
Federated backdoor attacks can preserve high ordinary accuracy while inducing targeted behavior on triggered inputs \cite{Bagdasaryan2020}; ordinary test accuracy therefore cannot by itself characterize trigger-conditioned behavior. Calibration adds another dimension: expected calibration error (ECE) summarizes agreement between confidence and correctness \cite{Guo2017}, but a low-accuracy model can still exhibit a numerically small ECE under some confidence distributions. Likewise, a minimum client accuracy is a lower-tail descriptive outcome rather than a complete fairness criterion; fairness-oriented FL work motivates richer distributional evaluation \cite{Li2020Fairness}. These distinctions motivate the present separation between ordinary accuracy, TTLR, ECE, and minimum-client accuracy, as well as the explicit separation of numerical summaries from execution and implementation provenance.

\section{Methods}

\subsection*{Evidence scope and terminology}
A \emph{task configuration} is one dataset--architecture pair, while \emph{canonical cell} is one method--dataset--architecture--condition--seed identity retained in the reconstructed matrix. Every canonical identity uses seed 1. The 25 task configurations therefore provide benchmark breadth but are not stochastic repetitions of one experiment. A \emph{run-summary record} is the retained machine-readable record containing final test accuracy and additional final metrics. A \emph{successful execution log} is stronger execution provenance because it documents a completed run. The manuscript distinguishes direct benchmark measurements, derived descriptive statistics, execution provenance, source-code observations, interpretation, and future design proposals throughout.

\subsection*{Federated-learning formulation}

Let client $i$ hold local dataset $D_i$ and empirical objective
\begin{equation}
F_i(\theta)=\frac{1}{|D_i|}\sum_{(x,y)\in D_i}\ell(f_\theta(x),y),
\label{eq:local}
\end{equation}
where $\theta$ denotes model parameters, $f_\theta$ the predictor, and $\ell$ the training loss. A population objective may be written
\begin{equation}
F(\theta)=\sum_{i=1}^{N}p_iF_i(\theta),\qquad p_i\ge 0,\qquad \sum_{i=1}^{N}p_i=1.
\label{eq:global}
\end{equation}
At communication round $t$, local training produces $\theta_{i,\mathrm{loc}}^t$ and update
\begin{equation}
\Delta_i^t=\theta_{i,\mathrm{loc}}^t-\theta_t.
\label{eq:update}
\end{equation}
Because the evaluated aggregators are not all representable by one scalar client-weight vector (for example, Trimmed Mean is coordinate-wise), we write the server update generally as
\begin{equation}
\theta_{t+1}=\theta_t+\eta_t\,\mathcal{A}_t\!\left(\{\Delta_i^t:i\in S_t\}\right),
\label{eq:generic}
\end{equation}
where $S_t$ is the selected-client set, $\eta_t$ is a server step factor, and $\mathcal{A}_t$ denotes the aggregation operator.

\subsection*{Aggregation methods and audited implementation behavior}
\subsubsection*{FedAvg.}
The sample-weighted form is
\begin{equation}
\mathcal{A}_{\mathrm{FedAvg}}=\sum_{i\in S_t}\frac{n_i}{\sum_{j\in S_t}n_j}\Delta_i^t,
\label{eq:fedavg}
\end{equation}
with $n_i=|D_i|$. The supplied scaffold's \texttt{fedavg\_aggregate} function uses these sample-count weights. This is a code-level observation for the audited scaffold; immutable source linkage is not available for every historical canonical cell.

\subsubsection*{Trimmed Mean.}
For coordinate $k$, let $\Delta_{(1),k}\le\cdots\le\Delta_{(m),k}$ be the sorted submitted values from $m=|S_t|$ clients. If $b$ values are removed from each end,
\begin{equation}
\left[\mathcal{A}_{\mathrm{TM}}\right]_k=\frac{1}{m-2b}\sum_{j=b+1}^{m-b}\Delta_{(j),k},\qquad 0\le b<\frac{m}{2}.
\label{eq:trim}
\end{equation}
The audited scaffold implements per-coordinate sorting with configurable \texttt{trim\_ratio}; its default is 0.2, so a ten-client round would trim two values from each tail. That setting is an implementation-level property of the supplied scaffold and is not asserted as a globally provenance-linked parameter for all canonical Trimmed Mean cells.

\subsubsection*{Krum.}
For a configured Byzantine bound $f$, Krum scores candidate update $i$ by
\begin{equation}
s_i=\sum_{j\in\mathcal{N}_i}\lVert\Delta_i^t-\Delta_j^t\rVert_2^2,
\qquad |\mathcal{N}_i|=m-f-2,
\label{eq:krumscore}
\end{equation}
and selects
\begin{equation}
i^\star=\arg\min_i s_i.
\label{eq:krumselect}
\end{equation}
The given scaffold implements single-update Krum, not Multi-Krum. It generates $f$ from the specified harmful fraction and the number of chosen clients. In the repaired subset, the maintained configuration is 10 chosen clients and malicious fraction 0.25; the audited Python path evaluates \texttt{round(0.25\,$\times$\,10)} to $f=2$. This value is not extended to canonical cells lacking the same configuration provenance, and no claim is made that Krum's theoretical assumptions were tested round-by-round for the whole matrix.

\subsubsection*{FLTrust.}
Let $\Delta_r^t$ denote a server root update. The directional trust score is
\begin{equation}
t_i=\max\!\left(0,\frac{\langle\Delta_i^t,\Delta_r^t\rangle}{\lVert\Delta_i^t\rVert_2\lVert\Delta_r^t\rVert_2}\right).
\label{eq:fltrust}
\end{equation}
In the given scaffold, each client update is rescaled to the root-update norm before trust-weighted averaging; negative cosine similarity is cut to zero, and if all trust scores are non-positive the implementation falls back to uniform weights. The root update is taught server-side using the same SGD learning rate, momentum, and weight decay parameters as the federated setup. These facts describe the audited implementation rather than providing entire historical history for every canonical FLTrust execution.

\subsubsection*{FedPARETO as implemented.}
The executable scaffold constructs four raw signals. Let $a_i$ denote anchor accuracy, $e_i$ anchor ECE, $\ell_i$ local client accuracy, $q_{i,t-1}$ temporal reliability, and $g_t$ the server root-update vector:
\begin{subequations}
\begin{align}
u_{i,\mathrm{acc}} &= a_i, \\
u_{i,\mathrm{cal}} &= -e_i, \\
u_{i,\mathrm{prio}} &= 1-\ell_i, \\
u_{i,\mathrm{rob}} &= \tfrac{1}{2}\max(0,\cos(\Delta_i^t,g_t))+\tfrac{1}{2}q_{i,t-1}.
\end{align}
\label{eq:fedpareto_components}
\end{subequations}
The term $1-\ell_i$ is described here as a \emph{low-local-accuracy priority signal}, not as a measured fairness outcome. Each component is min--max normalized across participating clients; the audited helper returns zero for an entire component if all participating values are equal. If $\widetilde{\mathbf{u}}_i$ is the normalized vector, the dominance-count rank is
\begin{equation}
r_i=1+\sum_{j\neq i}\mathbb{I}\!\left[\widetilde{\mathbf{u}}_j\succeq\widetilde{\mathbf{u}}_i\land \widetilde{\mathbf{u}}_j\neq\widetilde{\mathbf{u}}_i\right].
\label{eq:paretorank}
\end{equation}
The scalar utility used by the executable implementation is
\begin{equation}
s_i=\sum_k\alpha_k\widetilde{u}_{ik}+\frac{\lambda_P}{r_i}-\lambda_T(1-\widetilde{u}_{i,\mathrm{rob}}),
\label{eq:scalar}
\end{equation}
followed by
\begin{equation}
v_i=\exp(\tau s_i),
\label{eq:expweight}
\end{equation}

Euclidean projection of $\mathbf{v}$ onto the probability simplex, and uniform mixing when \texttt{entropy\_reg}$>0$. The verified source additionally initializes unseen-client temporal reliability to 0.5 and updates it as a momentum average of a signal combining the current aggregation weight and normalized robustness score. Legacy configuration files instantiate these values numerically, however the legacy settings are not regarded as canonical-wide hyperparameters since comprehensive per-run configuration snapshots are unavailable for the 500-cell matrix.

The executable approach is thus a \emph{Pareto-inspired dominance-count/scalarization heuristic}. It does not implement the conceptual draft's weighted-Tchebycheff optimization, explicit $\ell_1$ trust region around sample-size weights, reliability-dependent weight caps, benign-subspace singular-vector model, or formal constrained Pareto solver.

\subsection*{FedPARETO signal ownership and trust boundary}
The audited scaffold does not treat all utility signals equivalently. Table~\ref{tab:signal_ownership} records where each signal is produced. Anchor predictions and local metrics are computed in the client object and returned to the server; root-direction similarity and temporal reliability are server-computed. Consequently, the audited scaffold assumes integrity of client-returned predictive summaries unless an external trusted-execution or verification mechanism is added. No adaptive experiment in the retained benchmark tests deliberate forgery of these summary fields.

\begin{table}[ht]
\centering
\caption{\textbf{Signal ownership in the supplied FedPARETO scaffold.} The table describes audited source behavior, not immutable generating provenance for every historical benchmark cell.}
\label{tab:signal_ownership}
\begin{tabularx}{\linewidth}{p{0.24\linewidth}p{0.20\linewidth}X}
\toprule
Signal & Audited location & Interpretation / trust implication \\
\midrule
Anchor predictions / derived accuracy and ECE & Client computes; server consumes returned summary & Client-derived predictive evidence; forgeability is not evaluated in the benchmark \\
Local accuracy $\ell_i$ & Client computes and returns & Used only as low-local-accuracy priority; not an independently verified fairness measure \\
Root-direction similarity & Server computes from submitted delta and root vector & Directly tied to the submitted update in the audited server path \\
Temporal reliability $q_{i,t-1}$ & Server-maintained state & Initialized to 0.5 for unseen clients and updated by momentum from current weight and robustness score \\
\bottomrule
\end{tabularx}
\end{table}

\subsection*{Scope of the implementation audit}
The supplied FedPARETO scaffold is informative for mechanism audit but is not a complete generating repository for the full canonical benchmark. In particular, the audited scaffold's dataset/model registry does not instantiate all five canonical datasets and all five canonical architectures. The source audit therefore establishes behavior of the available implementation paths, including metric semantics and the update--utility mismatch described below, but it does not prove that every historical canonical cell was generated by exactly that revision or preprocessing/model-adaptation code.

\subsection*{Recorded threat conditions and configuration provenance}

The canonical summaries retain the labels Clean, Sign flipping, Gaussian, and BadNets for all corresponding cells. Exact attack strength and participation parameters are not uniformly provenance-linked across all 500 cells. Table~\ref{tab:threat_scope} therefore separates the global recorded label from subset-verified and code-level details.

```latex
\begin{table*}[t]
\centering
\caption{\textbf{Threat-condition evidence scope.}
Parameters in the third column are reported only at the evidence level at which they are verified; they are not generalized to all canonical cells.}
\label{tab:threat_scope}

\small
\setlength{\tabcolsep}{4pt}
\renewcommand{\arraystretch}{0.92}

\begin{tabularx}{\textwidth}{
    p{0.13\textwidth}
    p{0.24\textwidth}
    p{0.30\textwidth}
    X
}
\toprule
Condition &
Canonical evidence &
Audited / subset-verified detail &
Not established globally \\
\midrule

Clean &
Clean condition label retained for all clean cells &
No attack applied in the audited clean path &
Complete per-run configuration snapshot for every historical cell \\

Sign flipping &
Condition label retained for all sign-flipping cells &
Audited scaffold applies $-3\Delta$; repaired subset records malicious fraction 0.25 &
Global sign multiplier, malicious fraction, realized malicious count per round, and identical sampling schedule across all historical cells \\

Gaussian &
Condition label retained for all Gaussian cells &
Repaired subset records $\sigma=0.5$ and malicious fraction 0.25 &
Global $\sigma$, malicious fraction, realized malicious participation, and identical attack implementation across all historical cells \\

BadNets &
Condition label and retained target-label field available for all 125 BadNets cells &
Repaired subset records 30\% poisoned malicious batches, target 0, $4\times4$ trigger; audited metric triggers every test input &
Global target/trigger/poison fraction lineage and conventional target-excluding ASR \\

Adaptive / defense-aware attacks &
Not present in canonical matrix &
None &
Krum-aware, FLTrust-aware, FedPARETO-aware, model-replacement, and attack-strength sweeps \\

\bottomrule
\end{tabularx}
\end{table*}
```

The audited scaffold represents sign flipping as $\widetilde{\Delta}_i^t=-\lambda\Delta_i^t$ and Gaussian corruption as $\widetilde{\Delta}_i^t=\Delta_i^t+\varepsilon_i$ with $\varepsilon_i\sim\mathcal{N}(0,\sigma^2I)$. These equations describe the source implementation family; the benchmark-level analysis treats the historical labels categorically where exact per-cell parameters are unresolved.

\subsection*{Outcome measures}
The run summaries retain final global test accuracy, final ECE, and final worst-client accuracy for every canonical cell. For BadNets, the retained target-label field is additionally available. Configuration-matched attack-minus-clean change is
\begin{equation}
D_{m,d,h,a}=100\left(A_{m,d,h,a}-A_{m,d,h,\mathrm{clean}}\right),
\label{eq:delta}
\end{equation}
where $A$ is final accuracy, $m$ method, $d$ dataset, $h$ architecture, and $a$ an attacked condition. These are descriptive configuration-matched differences; a shared seed label does not show that the clean and attacked executions share all underlying stochastic trajectories, thus $D$ is not regarded as a seed-paired causal treatment effect.

For confidence-based calibration, let $B$ confidence bins divide the predictions, $I_b$ signify the samples in bin $b$, $\operatorname{acc}(I_b)$ their empirical accuracy, and $\operatorname{conf}(I_b)$ their mean confidence. ECE is
\begin{equation}
\mathrm{ECE}=\sum_{b=1}^{B}\frac{|I_b|}{N}\left|\operatorname{acc}(I_b)-\operatorname{conf}(I_b)\right|.
\label{eq:ece}
\end{equation}

The audited implementation utilises equal-width top-label confidence bins and defaults to $B=15$; the repaired subset specifically records 15 bins. This bin count is not stated as unchanging provenance for every historical canonical run. ECE is regarded exclusively as a calibration descriptor, not as a robustness score.

Worst-client accuracy is the minimum of the per-client evaluation accuracies retained by the evaluation path,
\begin{equation}
A_{\mathrm{worst}}=\min_{i\in\mathcal{C}_{\mathrm{eval}}} A_i.
\label{eq:worst}
\end{equation}
It is therefore a lower-tail descriptive outcome. It is not treated as a formal fairness guarantee, and the historical summaries do not uniformly preserve enough client-level sample-size context to make stronger fairness claims.

\subsection*{BadNets metric semantics: TTLR versus conventional ASR}
The supplied metric implementation applies the trigger to every test input and counts the fraction predicted as target label $y^\star$. We therefore denote the retained field
\begin{equation}
\mathrm{TTLR}=\frac{1}{N}\sum_{j=1}^{N}\mathbb{I}\!\left[f(T(x_j))=y^\star\right],
\label{eq:ttlr}
\end{equation}
where $T(\cdot)$ applies the trigger. Because Eq.~\eqref{eq:ttlr} includes examples whose natural label already equals $y^\star$, it differs from conventional target-excluding ASR,
\begin{equation}
\mathrm{ASR}_{\neg y^\star}=\frac{\sum_{j:y_j\neq y^\star}\mathbb{I}[f(T(x_j))=y^\star]}{\sum_{j:y_j\neq y^\star}1}.
\label{eq:asr}
\end{equation}
The retained aggregate summaries do not contain the class-conditioned triggered counts needed to reconstruct Eq.~\eqref{eq:asr}. Accordingly, the manuscript compares TTLR only as the observed all-triggered-input target-label frequency under the retained metric; lower TTLR is not translated into a general ranking of backdoor defenses.

\subsection*{Update--utility correspondence in the audited FedPARETO path}
In the audited sign-flipping/Gaussian client path, local training first creates an uncorrupted model and delta. The delta may then be modified by the attack, while local metrics and anchor predictions are subsequently computed from the still-uncorrupted local model object. Therefore
\begin{equation}
\Delta_{i,\mathrm{submitted}}^t\neq\theta_{i,\mathrm{summary}}^t-\theta_t
\label{eq:mismatch}
\end{equation}
can hold. Equation~\eqref{eq:mismatch} is a source-code observation: predictive evidence and submitted update can refer to different model objects. It does not establish that this mismatch caused the canonical FedPARETO accuracy reductions, nor that every canonical FedPARETO cell was generated by the exact audited revision.

A possible future update-consistent construction would evaluate the reconstructed candidate $\widehat{\theta}_i^t=\theta_t+\Delta_{i,\mathrm{submitted}}^t$ before calculating predictive utility. This construction is proposed only as a design direction; it is not experimentally evaluated in the present evidence and is not treated as an empirical contribution.

\section{Experimental setup and reconstruction protocol}

\subsection*{Canonical matrix}
The reconstructed matrix crosses five datasets (GTSRB, SVHN, MNIST, CIFAR-10, and CIFAR-100), five architecture labels (SimpleCNN, ResNet-18, MobileNetV3-Small, EfficientNet-B0, and ShuffleNetV2), five aggregation methods, and four recorded conditions, yielding 500 method--dataset--architecture--condition cells. All canonical identities encode seed 1. Exact architecture-adaptation and preprocessing details are not uniformly recoverable from a generating repository for every cell; the cross-architecture results are therefore treated as numerical comparisons across the recorded architecture labels rather than as a fully reproducible architecture-ablation study.

\subsection*{Canonicalization and execution provenance}
Canonical identity was defined by dataset, architecture, method, condition, and seed. Byte-identical archive copies were counted once for independent-evidence purposes. Numerical values were taken from the run-summary evidence associated with the canonical identity and cross-checked against the supplied rounded workbook; all 500 workbook final-accuracy entries agree with the canonical summaries within $5\times10^{-5}$. Provenance classification then used the strongest located execution evidence for each identity: an original successful execution log when located, otherwise a successful repaired/rerun execution log, and otherwise the run summary alone. No missing numerical value was imputed, and the provenance class was retained explicitly rather than used to select favorable numerical outcomes.

All 500 canonical cells have summary values. Successful execution logs were located for 490: 454 original and 36 repaired/rerun. The remaining 10 cells are Clean SVHN results for MobileNetV3-Small and ShuffleNetV2 across all five methods. They have summary records and workbook values but no successful execution log located in the supplied bundles. Table~\ref{tab:provenance} distinguishes numerical completeness from execution-log coverage.

\begin{table}[ht]
\centering
\caption{\textbf{Canonical evidence coverage and provenance.}}
\label{tab:provenance}
\begin{tabular}{lrr}
\toprule
Evidence category & Cells & Fraction \\
\midrule
Original execution log + summary & 454 & 90.8\% \\
Successful repaired/rerun log + summary & 36 & 7.2\% \\
Summary record & 10 & 2.0\% \\
\midrule
Total & 500 & 100\% \\
\bottomrule
\end{tabular}
\end{table}

The original suite master logs identify seed 1, a 500-round maximum, and early stopping disabled for their entries. The final repaired configurations independently record 500 rounds and, for that 36-run subset, 20 clients, 10 selected clients per round, one local epoch, batch size 64, learning rate 0.003, momentum 0.9, weight decay $5\times10^{-4}$, Dirichlet $\alpha=1.0$, malicious fraction 0.25 for attacked runs, anchor size 4096, and 15 ECE bins. These are subset-verified settings only.

\subsection*{Descriptive analysis and post-hoc provenance sensitivity}
Within each condition, methods were ranked separately within each dataset--architecture task; exact ties receive average rank and tied leaders divide one configuration win. We report macro means, medians, cross-task sample SD, IQR/range where useful, within-task ranks, configuration-win counts, configuration-matched attack-minus-clean differences, pairwise differences, and leave-one-dataset-out / leave-one-architecture-out sensitivity. These statistics describe the fixed observed task matrix and are not stochastic uncertainty estimates.

No $p$-values, t-tests, ANOVA, seed-level standard errors, or seed-level confidence intervals are reported for the canonical matrix because there is one identified seed per cell. The separate legacy/scaffold MNIST three-seed families are summarized only as supplementary evidence and are not used to create uncertainty intervals for the canonical matrix.

Two post-hoc provenance sensitivity analyses use only existing observations. First, the two task pairs whose clean cells are summary-only (SVHN--MobileNetV3-Small and SVHN--ShuffleNetV2) are removed in their entirety, yielding a matched 23-task / 460-cell subset. Second, we restrict to the 21 dataset--architecture task pairs for which all 20 method--condition cells have original successful execution logs, yielding an all-original 420-cell complete-task subset. The latter additionally excludes SVHN--ResNet-18 and CIFAR-100--ResNet-18, which contain repaired cells. These sensitivity analyses test dependence on provenance strata; they do not create new stochastic replication.

\section{Results}

\subsection*{Numerical reconstruction and provenance}
The reconstruction resolves the earlier incomplete 464/500 matrix by recovering numerical run-summary values for all 500 canonical identities. The workbook cross-check is exact after the documented rounding tolerance. The stronger provenance distinction remains important: numerical completeness is 100\%, whereas successful execution-log coverage is 98\%. All subsequent full-matrix analyses therefore use the 500 numerical summaries, while provenance sensitivity is reported separately rather than treating original, repaired, and summary-only evidence as identical.

\subsection*{Condition-level predictive performance}
Table~\ref{tab:summary} summarizes final global test accuracy across the fixed 25 task configurations. Under Clean, Trimmed Mean has the highest macro-mean accuracy (76.02\%) and lowest mean within-task rank (1.70), followed by FedAvg (70.92\%, rank 1.82). Under Sign flipping, Krum has the highest mean accuracy (63.91\%), lowest mean rank (1.32), and 19 configuration wins. Under Gaussian, Krum has mean accuracy 64.45\%, mean rank 1.08, and 24 wins. The BadNets ordinary-accuracy ordering returns to a pattern closer to Clean, with Trimmed Mean having the highest macro-mean ordinary accuracy (75.17\%) and FedAvg and Trimmed Mean tied at mean rank 1.80.

\begin{figure*}[ht]
\centering
\includegraphics[width=0.96\textwidth]{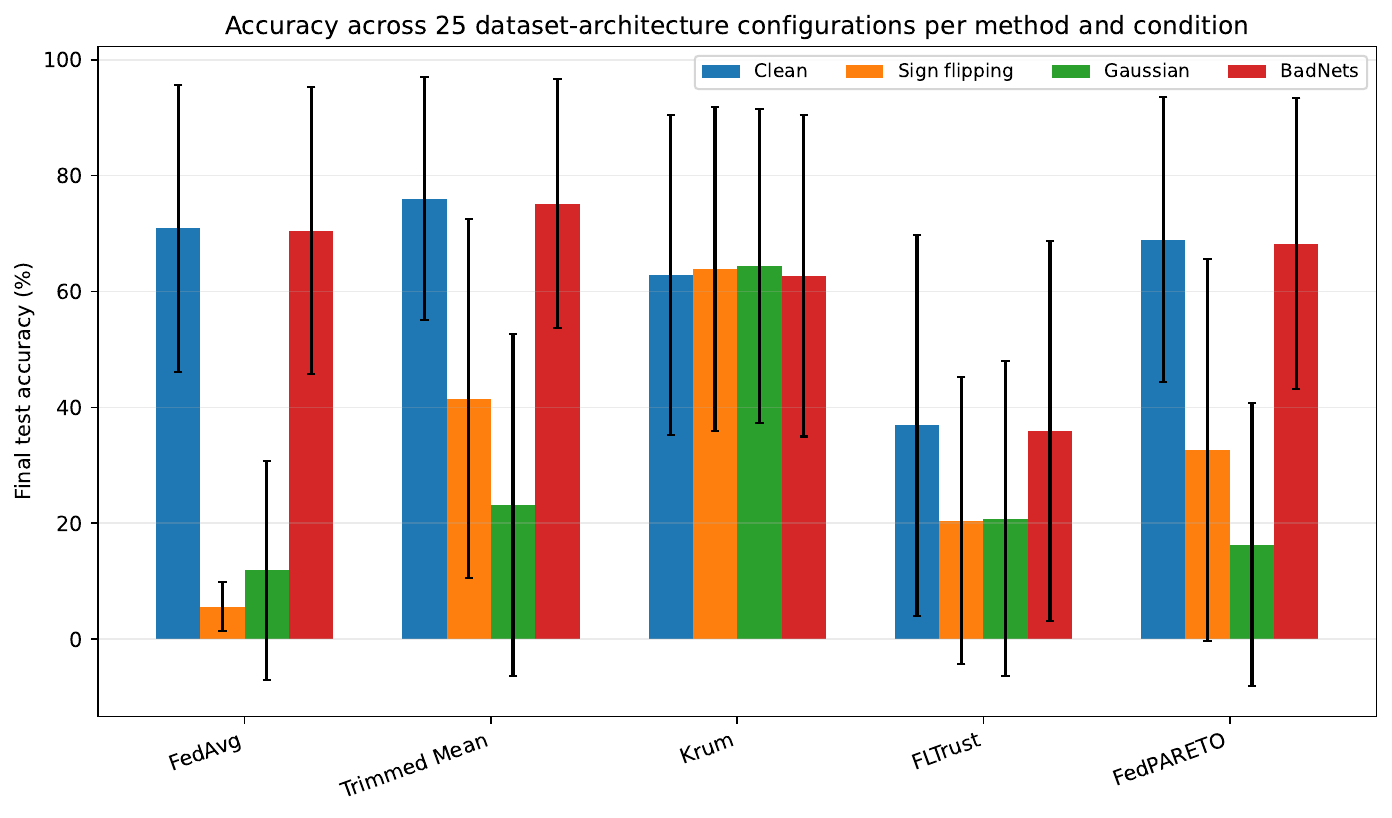}
\caption{\textbf{Final test accuracy across the reconstructed canonical matrix.} Bars show the macro mean across the 25 dataset--architecture task configurations for each method and recorded condition. Whiskers show cross-task sample SD, i.e. dispersion across heterogeneous tasks rather than uncertainty around a stochastic mean. The figure is descriptive and should be read together with within-task ranks and task-level distributions.}
\label{fig:accuracy}
\end{figure*}

\begin{table*}[ht]
\centering
\caption{\textbf{Condition-level final-accuracy summaries across the 25 task configurations.} SD is sample dispersion across heterogeneous task configurations, not seed-level uncertainty. Fractional wins divide a task-level win among tied leaders.}
\label{tab:summary}
\resizebox{\textwidth}{!}{%
\begin{tabular}{llrrrrrr}
\toprule
Condition & Method & Mean (\%) & Median (\%) & SD (pp) & Mean rank & Wins & $n$ tasks \\
\midrule
Clean & FedAvg & 70.92 & 75.39 & 24.78 & 1.82 & 14.5 & 25 \\
& Trimmed Mean & 76.02 & 85.95 & 21.01 & 1.70 & 10.0 & 25 \\
& Krum & 62.85 & 72.17 & 27.58 & 3.80 & 0.0 & 25 \\
& FLTrust & 36.88 & 24.15 & 32.81 & 4.72 & 0.0 & 25 \\
& FedPARETO & 68.93 & 72.66 & 24.61 & 2.96 & 0.5 & 25 \\
\addlinespace
Sign flipping & FedAvg & 5.59 & 6.70 & 4.22 & 4.44 & 0.0 & 25 \\
& Trimmed Mean & 41.48 & 45.11 & 30.97 & 2.78 & 1.0 & 25 \\
& Krum & 63.91 & 72.37 & 27.98 & 1.32 & 19.0 & 25 \\
& FLTrust & 20.47 & 11.06 & 24.83 & 3.56 & 0.0 & 25 \\
& FedPARETO & 32.63 & 10.54 & 32.90 & 2.90 & 5.0 & 25 \\
\addlinespace
Gaussian & FedAvg & 11.88 & 6.70 & 18.91 & 4.04 & 0.0 & 25 \\
& Trimmed Mean & 23.14 & 9.80 & 29.56 & 3.14 & 1.0 & 25 \\
& Krum & 64.45 & 72.70 & 27.10 & 1.08 & 24.0 & 25 \\
& FLTrust & 20.82 & 10.00 & 27.23 & 3.02 & 0.0 & 25 \\
& FedPARETO & 16.31 & 9.80 & 24.38 & 3.72 & 0.0 & 25 \\
\addlinespace
BadNets & FedAvg & 70.51 & 76.64 & 24.79 & 1.80 & 14.0 & 25 \\
& Trimmed Mean & 75.17 & 85.29 & 21.49 & 1.80 & 8.0 & 25 \\
& Krum & 62.75 & 68.56 & 27.77 & 3.80 & 0.0 & 25 \\
& FLTrust & 35.99 & 21.99 & 32.79 & 4.72 & 0.0 & 25 \\
& FedPARETO & 68.29 & 72.68 & 25.05 & 2.88 & 3.0 & 25 \\
\bottomrule
\end{tabular}}
\end{table*}

\begin{figure}[ht]
\centering
\includegraphics[width=0.80\linewidth]{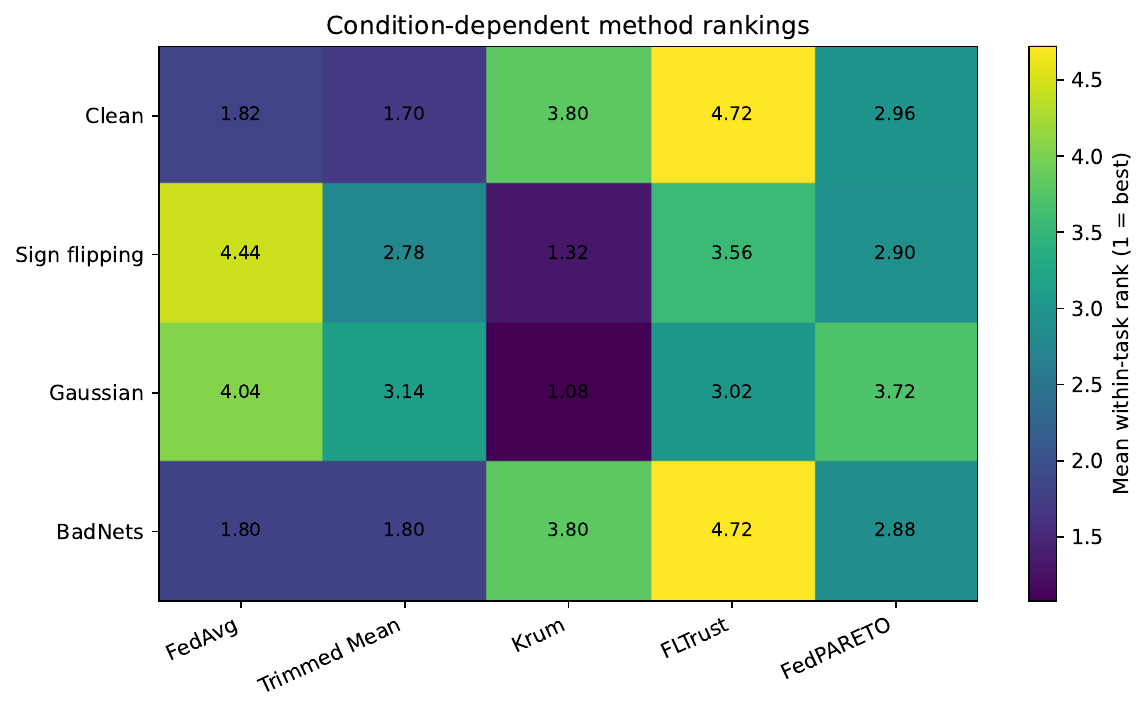}
\caption{\textbf{Mean within-task ranks across recorded conditions.} Each dataset--architecture task is ranked separately, with rank 1 denoting the highest final accuracy among the five methods. The clean and BadNets ordinary-accuracy ordering differs from the Krum-first ordering under Sign flipping and Gaussian.}
\label{fig:ranks}
\end{figure}

\begin{figure*}[ht]
\centering
\includegraphics[width=0.97\textwidth]{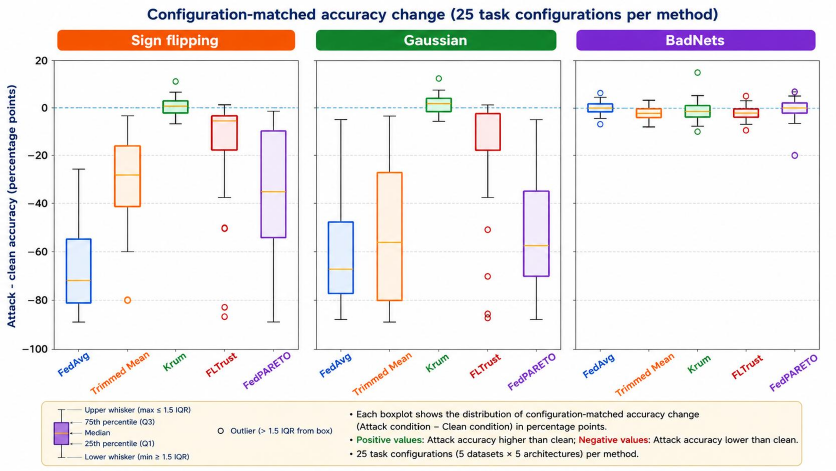}
\caption{\textbf{Configuration-matched attack-minus-clean accuracy changes.} Task-level percentage-point differences for a single technique are included in each distribution. Method, dataset, and architecture are used for matching. The distributions are descriptive rather than seed-paired treatment effects since a shared seed label does not produce identical stochastic trajectories across clean and attacked runs.}
\label{fig:deltas}
\end{figure*}

Instead of aggregating absolute task difficulty, Figure~\ref{fig:ranks} highlights relative method ranking inside a challenge. The rank view is complemented by configuration-matched attack-minus-clean adjustments in Figure~\ref{fig:deltas}. These are not distinct replications, but rather many descriptions of the same fixed runs.

The mean matched change under sign flipping is $-65.32$ pp for FedAvg, $-34.54$ pp for Trimmed Mean, $+1.06$ pp for Krum, $-16.41$ pp for FLTrust, and $-36.30$ pp for FedPARETO. The comparable changes in Gaussian are $-59.04$, $-52.88$, $+1.60$, $-16.07$, and $-52.62$ pp. The tiny positive Krum means are not understood as attack-induced improvement; rather, the supporting finding is that Krum's reported attacked accuracies are similar to its clean accuracies in the seed-1 matrix.

\subsection*{Provenance sensitivity of the headline ordering}
Simply stating that the 10 summary-only cells make up 2\% of cells understates their downstream importance in matched clean-reference studies since they are concentrated in two clean SVHN task pairings. Table~\ref{tab:prov_sensitivity} therefore compares the full matrix with a 23-task complete-case subset excluding those task pairs and with a stricter 21-task subset in which every method--condition cell has an original successful execution log.

\begin{table*}[ht]
\centering
\caption{\textbf{Post-hoc provenance sensitivity using existing data only.} The 23-task set removes the two task pairs whose clean reference is summary-only. The 21-task set retains only task pairs for which all 20 method--condition cells have original successful logs. Values are mean within-task rank followed by configuration wins for the leading method.}
\label{tab:prov_sensitivity}
\scriptsize
\resizebox{\textwidth}{!}{%
\begin{tabular}{lcllll}
\toprule
Analysis set & Tasks & Clean & Sign flipping & Gaussian & BadNets ordinary \\
\midrule
Full canonical matrix & 25 & Trimmed Mean 1.70 / 10 & Krum 1.32 / 19 & Krum 1.08 / 24 & FedAvg 1.80 / 14; TM 1.80 / 8 \\
Exclude summary-only clean task pairs & 23 & Trimmed Mean 1.76 / 8 & Krum 1.30 / 18 & Krum 1.09 / 22 & FedAvg 1.83 / 13; TM 1.83 / 7 \\
All-original complete-task subset & 21 & Trimmed Mean 1.79 / 7 & Krum 1.33 / 16 & Krum 1.10 / 20 & FedAvg 1.86 / 12; TM 1.86 / 6 \\
\bottomrule
\end{tabular}}
\end{table*}

The Krum-first ordering under Sign flipping and Gaussian is retained in both provenance-sensitive subsets. Krum's mean matched attack-minus-clean change is $+0.84$ pp (Sign flipping) and $+1.58$ pp (Gaussian) in the 23-task set, and $+0.83$ pp and $+1.42$ pp in the 21-task all-original set, compared with $+1.06$ pp and $+1.60$ pp in the full matrix. This reanalysis does not establish stochastic repeatability; it shows that the headline fixed-matrix ordering is not created solely by the 10 summary-only cells or by the repaired/rerun task pairs.

\subsection*{Global and lower-tail client performance}
\begin{figure*}[ht]
\centering
\includegraphics[width=0.86\textwidth]{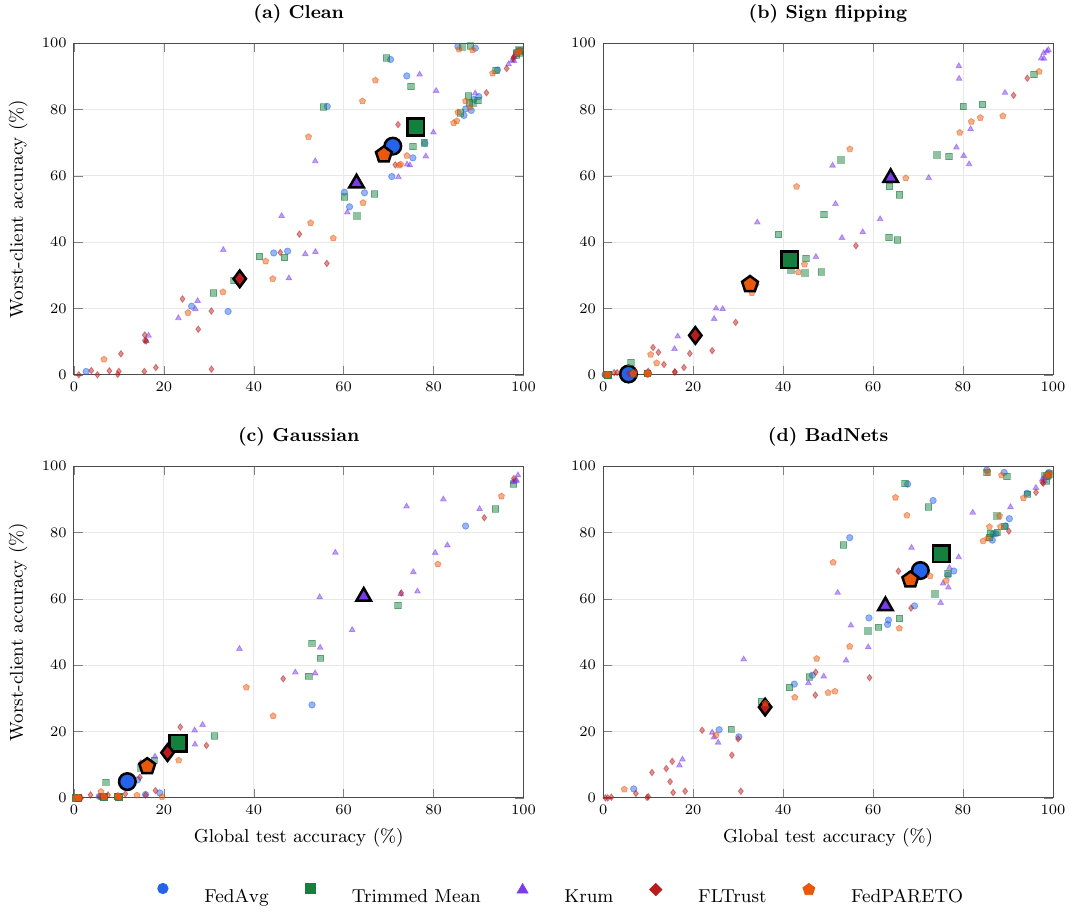}
\caption{\textbf{Global versus worst-client accuracy across the four recorded conditions.} Each small marker is one dataset--architecture task and the large outlined marker is the method mean. Worst-client accuracy is the minimum recorded client evaluation accuracy and is interpreted as a lower-tail descriptive outcome, not a fairness guarantee. The plotted task distributions are not seed-level uncertainty estimates.}
\label{fig:all4_accuracy_worst}
\end{figure*}

Figure~\ref{fig:all4_accuracy_worst} compares final global accuracy with the minimum recorded client accuracy. In Clean, mean worst-client accuracy is 74.72\% for Trimmed Mean, 68.94\% for FedAvg, 66.39\% for FedPARETO, 57.80\% for Krum, and 28.94\% for FLTrust. Under Sign flipping, the corresponding values are 34.70\%, 0.23\%, 27.28\%, 59.44\%, and 11.86\%; under Gaussian they are 16.50\%, 4.90\%, 9.49\%, 60.77\%, and 13.65\%, respectively. Thus the supplied summaries record comparatively high minimum-client accuracy for Krum under the two model-poisoning labels, but this outcome is not equated with formal fairness.

\subsection*{BadNets ordinary accuracy and retained triggered target-label behavior}
BadNets ordinary accuracy is numerically close to Clean accuracy at the macro-mean level: the mean matched changes are $-0.41$ pp for FedAvg, $-0.86$ pp for Trimmed Mean, $-0.10$ pp for Krum, $-0.89$ pp for FLTrust, and $-0.64$ pp for FedPARETO. This similarity concerns untriggered predictive performance only.

Under the supplied audited metric semantics, mean TTLR is 72.68\% for FedAvg, 69.70\% for Trimmed Mean, 60.54\% for FedPARETO, 30.61\% for Krum, and 24.41\% for FLTrust. Table~\ref{tab:badnets} separates ordinary accuracy, TTLR, ECE, and minimum-client accuracy because they answer different questions. In particular, the low mean TTLR of FLTrust occurs together with low ordinary accuracy, which illustrates why TTLR should not be translated directly into a backdoor-defense ranking.

\begin{table}[ht]
\centering
\caption{\textbf{BadNets performance across five datasets and 25 job settings.}
TTLR is not the conventional target-excluding attack success rate (ASR);
under the audited metric semantics, it denotes the frequency with which
triggered inputs are assigned the target label. Worst-client accuracy and
ECE are reported separately as descriptive metrics.}
\label{tab:badnets}
\scriptsize
\begin{tabular}{lrrrrrr}
\toprule
Method & GTSRB & SVHN & MNIST & CIFAR-10 & CIFAR-100 & Macro Mean \\
\midrule
FedAvg       & 92.37 & 73.39 & 68.59 & 55.80 & 73.26 & 72.68 \\
Trimmed Mean & 89.53 & 43.76 & 99.89 & 56.30 & 59.04 & 69.70 \\
Krum         & 15.38 & 26.76 & 63.91 & 45.37 &  1.64 & 30.61 \\
FLTrust      & 23.73 & 19.49 &  9.18 & 28.10 & 41.57 & 24.41 \\
FedPARETO    & 78.96 & 49.24 & 67.94 & 54.58 & 51.96 & 60.54 \\
\bottomrule
\end{tabular}
\end{table}

Cross-dataset TTLR averaging is additionally difficult to interpret because the datasets differ in class count and target-class prevalence, and target label 0 is provenance-verified only for the repaired subset. Table~\ref{tab:ttlr_dataset} therefore reports the existing 125 BadNets TTLR values after stratification by dataset. These are still retained-metric summaries, not conventional ASR.

\begin{table*}[ht]
\centering
\caption{\textbf{Dataset-stratified mean TTLR (\%) across the five architectures.} Values are descriptive summaries of the maintained all-triggered-input metric. Differences in dataset class structure, target prevalence, and insufficient global target-label provenance hinder reading this table as a traditional backdoor-defense scoreboard.}
\label{tab:ttlr_dataset}
\begin{tabular}{lrrrrrr}
\toprule
Method & GTSRB & SVHN & MNIST & CIFAR-10 & CIFAR-100 & Macro mean \\
\midrule
FedAvg & 92.37 & 73.39 & 68.59 & 55.80 & 73.26 & 72.68 \\
Trimmed Mean & 89.53 & 43.76 & 99.89 & 56.30 & 59.04 & 69.70 \\
Krum & 15.38 & 26.76 & 63.91 & 45.37 & 1.64 & 30.61 \\
FLTrust & 23.73 & 19.49 & 9.18 & 28.10 & 41.57 & 24.41 \\
FedPARETO & 78.96 & 49.24 & 67.94 & 54.58 & 51.96 & 60.54 \\
\bottomrule
\end{tabular}
\end{table*}

\subsection*{Matched FedPARETO--Krum comparison}
FedPARETO exceeds Krum in 22/25 Clean configurations with a mean FedPARETO-minus-Krum difference of $+6.08$ pp. Under Sign flipping it exceeds Krum in 5/25 configurations (mean $-31.28$ pp), and under Gaussian in 1/25 (mean $-48.15$ pp). Under BadNets ordinary accuracy, FedPARETO again exceeds Krum in 22/25 configurations (mean $+5.54$ pp). The same reversal remains in the 21-task all-original subset: mean differences are $+4.82$, $-32.21$, $-46.99$, and $+4.53$ pp for Clean, Sign flipping, Gaussian, and BadNets ordinary accuracy, respectively.

\begin{figure}[ht]
\centering
\includegraphics[width=0.62\linewidth]{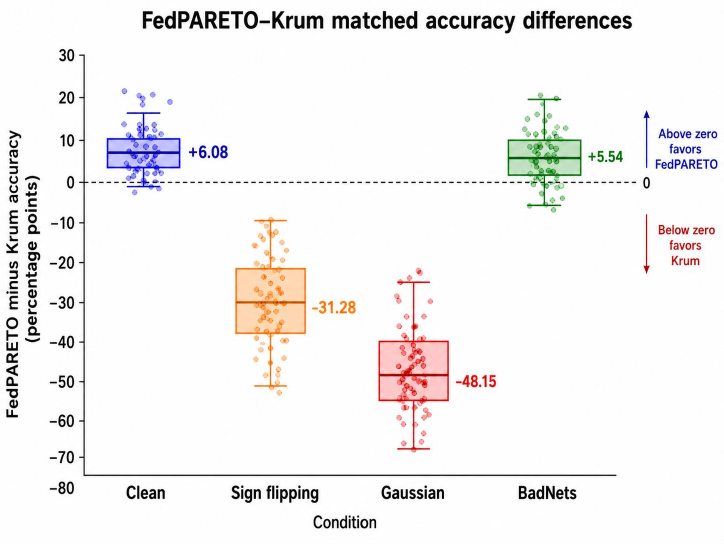}
\caption{\textbf{Matched FedPARETO--Krum final-accuracy differences.} Positive values favour FedPARETO and negative values favour Krum. Each point is a dataset--architecture task. The distributions are descriptive across fixed tasks and do not represent seed-level uncertainty or a causal comparison of the two design principles.}
\label{fig:paretokrum}
\end{figure}

Figure~\ref{fig:paretokrum} highlights this pair because it links the empirical ranking reversal to the FedPARETO implementation audit. A complete ordered pairwise method table is provided in the Supplementary Information so that the comparison is not treated as the only relevant pairwise result. The two methods also operate under different information assumptions: the supplied FedPARETO scaffold uses server/root and client-returned predictive information, whereas Krum uses submitted-update geometry.

\subsection*{Leave-one-group-out sensitivity and supplementary evidence}
Leaving out each dataset in turn leaves Krum at rank 1 under both Sign flipping and Gaussian in all five exclusions; its leave-one-dataset-out mean accuracy ranges from 55.34\% to 74.46\% for Sign flipping and 56.01\% to 74.73\% for Gaussian. Leaving out each architecture likewise leaves Krum at rank 1 in all five exclusions; the corresponding mean-accuracy ranges are 61.67--65.20\% and 62.21--65.73\%. These are descriptive sensitivity checks showing that the Krum-first ordering in the fixed matrix is not dependent on one named dataset or architecture group; they do not establish stochastic or external generalization. Full values are provided in Supplementary Figs. S7--S8 and their source tables.

Seven separate legacy/scaffold MNIST families contain seeds 1--3. Their seed-level SDs range from 0.08 to 7.14 pp, demonstrating that stochastic variability in this broader experimental family can be small or several percentage points depending on method and condition. These sparse, unbalanced families do not validate the canonical matrix and are not pooled with it. Additionally, anchor-size observations and single-run FedPARETO ablations are solely kept as exploratory supplemental evidence. The four recorded anchor-size points are just non-monotonic observations rather than a scaling rule, and the available ablations do not determine the marginal causal contribution of any utility component.

\subsection*{FedPARETO implementation audit}
Anchor accuracy, negative ECE, low-local-accuracy priority, root-direction similarity, temporal reliability, per-round min--max normalisation, dominance-count ranking, scalar scoring, exponential transformation, simplex projection, and uniform mixing are all implemented by the provided scaffold, according to the source audit. Several more comprehensive methods outlined in the conceptual draft are not implemented. The most significant source observation is Eq.~\eqref{eq:mismatch}: even tho the supplied delta is different, the client-returned prediction summaries may still represent the uncorrupted locally trained model after post-training sign-flipping or Gaussian corruption.

\begin{figure*}[ht]
\centering
\includegraphics[width=\textwidth]{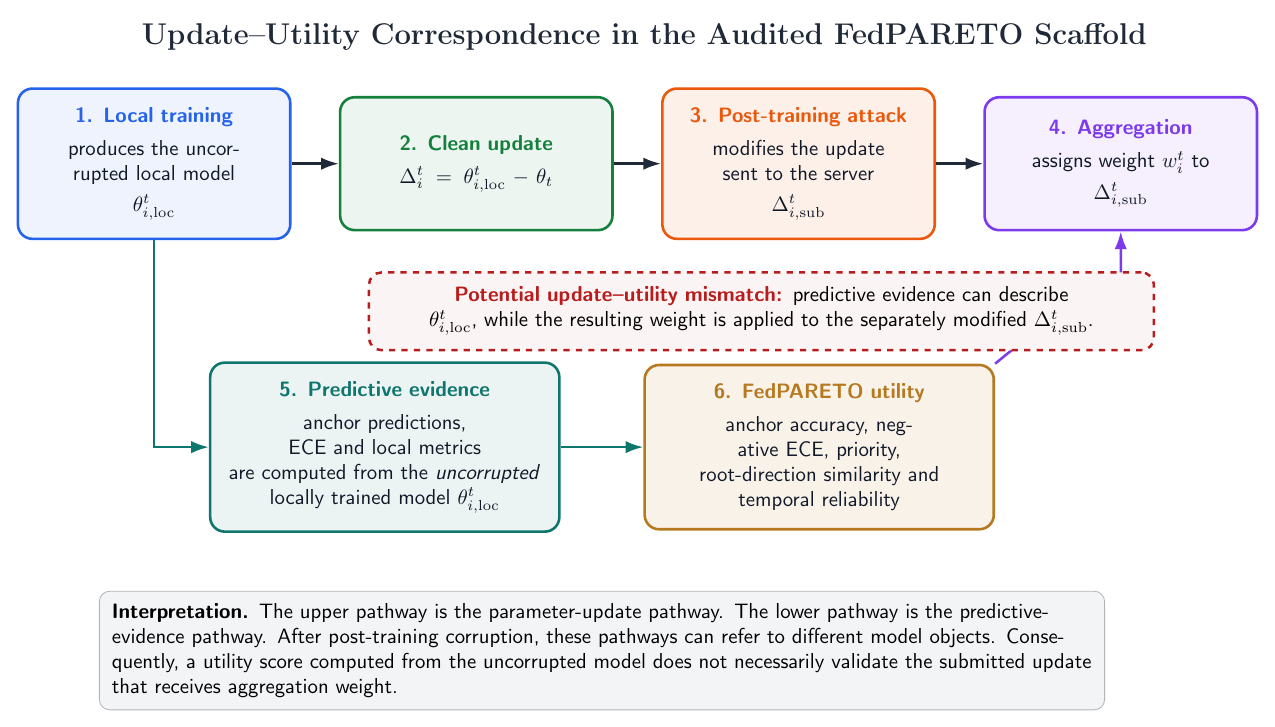}
\caption{\textbf{Update--utility correspondence in the audited FedPARETO scaffold.} The upper pathway represents the submitted parameter update; the lower pathway represents predictive evidence used in utility calculation. In the audited sign-flipping/Gaussian path, predictive evidence can remain associated with the uncorrupted locally trained model even after the submitted delta is modified. The figure documents a source-level failure mode and does not establish that the mismatch caused the canonical FedPARETO outcomes or that every historical cell used this exact revision.}
\label{fig:mismatch}
\end{figure*}

These two paths are separated in Figure~\ref{fig:mismatch}. This is not an empirically proven reason for the standard FedPARETO findings; rather, it is an implementation-level failure mode. The source-code mismatch and the empirical FedPARETO results are provided as separate types of evidence as the precise generating revision is not inextricably tied to each canonical cell.

\section{Discussion}

\subsection*{Method ordering differs across the recorded conditions}
The main empirical finding is that the five assessed approaches are arranged differently across the fixed seed-1 matrix. While Krum achieves the best final accuracy in 19/25 Sign-flipping and 24/25 Gaussian task designs, clean performance prefers FedAvg and Trimmed Mean by macro mean and within-task rank. A clean-like arrangement is restored by BadNets' ordinary accuracy. These findings of the reconstructed matrix are condition-specific and do not prove that one aggregator is better overall.

The provenance sensitivity analysis strengthens this narrow interpretation. Removing the two task pairs whose clean baselines are summary-only, and then restricting further to 21 task pairs for which every method--condition cell has an original successful log, does not change the Krum-first ordering under Sign flipping or Gaussian. This is useful evidence against a provenance-artifact explanation of the headline ordering, but it does not address seed-to-seed repeatability or unresolved historical attack-strength differences.

\subsection*{Interpreting the Krum result}
Krum's recorded performance is consistent with a regime in which its whole-update neighborhood-distance rule is well matched to the geometry of the evaluated sign-reversal and additive-noise conditions. That interpretation is plausible because conspicuous direction or distance deviations are exactly the information Krum uses. It is not a demonstrated causal mechanism for every task, and it should not be generalized to adaptive Byzantine adversaries. The benchmark does not contain Krum-aware optimization, model replacement, malicious-fraction sweeps, attack-strength sweeps, or controlled heterogeneity sweeps. Moreover, the configured Byzantine bound and realized malicious participation are not provenance-linked globally. The strongest defensible conclusion is therefore that Krum recorded the best final-accuracy ordering among the five evaluated methods under the retained Sign-flipping and Gaussian condition labels.

\subsection*{FedAvg, Trimmed Mean, and trust-based methods}
FedAvg's large negative matched changes under Sign flipping and Gaussian are consistent with an averaging rule that has no intrinsic rejection mechanism for arbitrary submitted updates. Trimmed Mean retains substantially higher recorded accuracy than FedAvg under Sign flipping in this matrix, but its Gaussian macro mean remains far below Krum's. These results illustrate that coordinate-wise trimming and whole-update geometric selection respond differently to the evaluated corruption patterns; they do not establish a general security hierarchy.

FLTrust and FedPARETO use additional trusted or validation information. This makes their operational assumptions different from those of Krum and Trimmed Mean. In the audited FedPARETO scaffold, anchor predictions and local accuracy are client-derived summaries, while root similarity and temporal reliability are server-computed. A stronger adversary able to forge client-returned summaries could therefore target the utility channel itself; such attacks were not evaluated. The evaluated implementations also require access to individual client updates or per-client statistics and are not directly compatible with a protocol that reveals only an aggregate update without additional secure protocol design. % [REFERENCE TO VERIFY: secure aggregation compatibility literature]

\subsection*{Update--utility correspondence is an implementation problem, not a causal outcome}

The FedPARETO code audit identifies a concrete software-level inconsistency: the object used to derive predictive utility can differ from the submitted update receiving aggregation weight. The empirical matrix concurrently demonstrates substantial FedPARETO reductions under the recorded Sign-flipping and Gaussian labelling. These two observations should not be collapsed into a causal explanation. The available evidence cannot isolate whether the mismatch, utility normalization, anchor representativeness, trust parameters, attack geometry, historical source differences, or combinations of these factors contributed to the recorded outcomes.

The general design implication is narrower: when predictive evidence is used to assign aggregation influence, the implementation should make explicit which model/update object that evidence validates. Reconstructing $\theta_t+\Delta_{i,\mathrm{submitted}}^t$ and evaluating that candidate server-side is one possible future update-consistent design. Its performance, calibration, cost, and security remain unevaluated here.

\subsection*{BadNets metric semantics and interpretation}
The BadNets results show why ordinary accuracy and trigger-conditioned target-label behavior must be separated. FedAvg and Trimmed Mean retain high ordinary accuracy while their mean TTLR values are near 70\%. Conversely, FLTrust has the lowest mean TTLR but also the lowest ordinary accuracy among the five methods. TTLR therefore records a real behavior of the retained metric but is not sufficient for a conventional backdoor-defense ranking.

The metric audit also imposes a provenance boundary. The supplied metric implementation clearly triggers every test input and therefore computes TTLR rather than target-excluding ASR, but an immutable metric-source hash is not linked to every historical BadNets result. We therefore report the historical field using the semantics established by the supplied audited implementation while explicitly noting that exact code-to-result lineage is incomplete. Conventional ASR, clean-model triggered baselines, and source-class-conditioned trigger behavior cannot be reconstructed from the retained aggregate summaries.

\subsection*{Reproducibility implications}
The reconstruction demonstrates why numerical completeness and execution reproducibility should be stated separately. The matrix is numerically complete at 500/500 summaries, but successful execution logs were located for 490/500 cells and immutable source/configuration lineage is not complete. Repaired executions also span different software environments. The final repaired bundle records Python 3.10.13 and PyTorch 2.5.1 on NVIDIA L40S hardware, whereas an earlier repair attempt records Python 3.13.5 and PyTorch 2.8.0+cu128 on the same hardware class. The final accuracies are retained as canonical summary values, but these heterogeneous environments are not treated as a controlled runtime experiment and no cross-method latency or throughput claim is made.

\section{Limitations and future work}
The canonical matrix contains one identified seed per cell. Consequently, it cannot estimate run-to-run variability, stochastic ranking probability, or statistical superiority. Cross-task SD, ranks, win counts, matched differences, leave-one-group-out results, and provenance sensitivity are descriptive analyses of the fixed observed configurations. The limited three-seed MNIST families do not repair this limitation.

Exact attack and training parameters are not provenance-linked uniformly across all historical cells. This prevents treating the retained condition labels as globally verified attack-strength treatments and prevents invoking theoretical robustness guarantees that require exact Byzantine participation. Additionally, the benchmark does not include a controlled non-IID severity sweep, model replacement, attack-strength and malicious-fraction sweeps, or adaptive defense-aware poisoning. As a consequence, the Krum result is limited to the five assessed techniques and the recorded threat categories. The provided audited scaffold is not the full generating repository for all five named datasets and five architectural labels included in the numerical matrix. Therefore, not every cell can completely recover from precise preprocessing, architectural adaption, and immutable source revision. The origin of ten clean SVHN cells is summary-only. Stronger reproducibility would need per-run source hashes, full configuration snapshots, data-split identities, environment locks, checkpoint hashes, and saved round-level logs, even tho the 21-task all-original sensitivity maintains the headline ordering.

TTLR is not a traditional target-excluding ASR. Additionally, source-class-conditioned backdoor analysis and clean-trigger baselines are not supported by the summaries that were kept.ECE is sensitive to confidence binning and can be numerically small in low-accuracy models; worst-client accuracy is an extreme lower-tail statistic without uniformly retained client-size context. These outcomes are therefore reported descriptively and not combined into a single robustness or fairness score.

The FedPARETO ablations and anchor-size observations are single-run legacy/scaffold experiments and do not establish component causality or an anchor-size response law. The proposed update-consistent FedPARETO construction is not experimentally evaluated. Future work would require balanced seeds, verified threat/configuration manifests, adaptive attacks, broader baseline coverage, conventional ASR, client-distribution analysis, and controlled computational-cost measurement before stronger robustness or deployment claims could be made.

\section{Conclusion}
A clear, constrained conclusion is backed by the reconstructed data. The ranking of the five assessed aggregation techniques varies significantly between the recorded circumstances in the fixed seed-1 matrix. In 19/25 Sign-flipping and 24/25 Gaussian task configurations, Krum achieves the greatest final accuracy, whereas Trimmed Mean has the highest clean macro-mean accuracy. After deleting the summary-only task pairs and restricting the analysis to 21 task pairs with original successful execution logs for each method--condition cell, the Krum-first ordering is still in situ. Both universal Byzantine robustness and stochastic repeatability are not proved by these descriptive facts.
Although the retained triggered target-label metric shows a different trend, BadNets' conventional accuracy often approaches clean accuracy. Since that field is TTLR rather than traditional target-excluding ASR under the provided certified metric implementation, it shouldn't be used by itself to rank backdoor defences. Predictive evidence and the weighted submitted update may relate to distinct model objects in an update--utility correspondence failure mode, according to the FedPARETO source audit. This is not a proven cause of the standard FedPARETO results; rather, it is an implementation observation. Overall, the analysis demonstrates that threat scope, metric semantics, implementation specifics, and provenance strength significantly influence comparative FL-security claims, and that these kinds of evidence have to be disclosed independently.

\subsection*{Acknowledgement}

This work was supported by the Variable Energy Cyclotron Centre (VECC), Department of Atomic Energy (DAE), Government of India (GoI), through the provision of GPU resources and computational facilities essential for carrying out this research. The authors gratefully acknowledge the technical support and computational infrastructure provided by VECC. The authors also sincerely appreciate the peer reviewers for their insightful comments and constructive criticism, which helped improve the quality of this work.

\subsection*{Author contributions statement}

S.M. conceived the study and developed the core methodology, led the
experimental evaluation, performed the model training and benchmarking
experiments, conducted the ablation analyses, prepared the corresponding
tables and figures, generated the manuscript visualizations, and contributed
substantially to manuscript writing and revision. V.K.R. developed the
benchmarking scripts for latency and throughput evaluation, assisted with
system profiling, and prepared the reproducibility resources. T.S. contributed
to model selection, reviewed the training and experimental configurations
with particular attention to hardware efficiency, and independently validated
the final results. All authors contributed to the critical review of the
manuscript, approved the final version, and agreed to be accountable for the
work.

\subsection*{Additional information}

\textbf{Accession codes}: Not applicable.\\
\textbf{Competing interests}: The authors declare no competing interests. The corresponding author is responsible for ensuring that this statement is accurate and has been agreed upon by all co-authors. All code, training logs, configuration files, and evaluation scripts used in this study are publicly available at~\url{https://github.com/mazumdarsoumya/RobustFL-Bench}.

\subsection*{Declarations}

\subsubsection*{Data availability}
All source code, configuration files, and supplementary scripts used in this study are publicly available at \url{https://github.com/mazumdarsoumya/RobustFL-Bench}. The datasets are available under its original license and cannot be redistributed by the authors. Due to institutional data-sharing restrictions, detailed training logs and key result files are not openly available but can be accessed upon reasonable request to the corresponding author at reachme@soumyamazumdar.com.

% \subsubsection*{Funding}
% The Department of Atomic Energy (DAE), Government of India (GoI), is also appreciated for financing the open-access publishing of this study.

\subsubsection*{Clinical Trial Number}
Clinical trial number: not applicable.

\subsubsection*{Ethics, Consent to Participate, and Consent to Publish}
Ethics, Consent to Participate, and Consent to Publish declarations: not applicable.

\subsubsection*{Conflict of Interest}
All authors assert that they own no financial or personal affiliations that may be seen as affecting the work provided in this study. No conflicts of interest are acknowledged.

\bibliographystyle{IEEEtran}
\bibliography{privfedtalk_refs}

@inproceedings{McMahan2017,
  author    = {McMahan, H. Brendan and Moore, Eider and Ramage, Daniel and Hampson, Seth and Ag{\"u}era y Arcas, Blaise},
  title     = {Communication-Efficient Learning of Deep Networks from Decentralized Data},
  booktitle = {Proceedings of the 20th International Conference on Artificial Intelligence and Statistics},
  series    = {Proceedings of Machine Learning Research},
  volume    = {54},
  pages     = {1273--1282},
  year      = {2017},
  publisher = {PMLR}
}

@inproceedings{Li2020FedProx,
  author    = {Li, Tian and Sahu, Anit Kumar and Zaheer, Manzil and Sanjabi, Maziar and Talwalkar, Ameet and Smith, Virginia},
  title     = {Federated Optimization in Heterogeneous Networks},
  booktitle = {Proceedings of Machine Learning and Systems},
  volume    = {2},
  pages     = {429--450},
  year      = {2020}
}

@inproceedings{Karimireddy2020SCAFFOLD,
  author    = {Karimireddy, Sai Praneeth and Kale, Satyen and Mohri, Mehryar and Reddi, Sashank J. and Stich, Sebastian U. and Suresh, Ananda Theertha},
  title     = {{SCAFFOLD}: Stochastic Controlled Averaging for Federated Learning},
  booktitle = {Proceedings of the 37th International Conference on Machine Learning},
  series    = {Proceedings of Machine Learning Research},
  volume    = {119},
  pages     = {5132--5143},
  year      = {2020},
  publisher = {PMLR}
}

@inproceedings{Wang2020FedNova,
  author    = {Wang, Jianyu and Liu, Qinghua and Liang, Hao and Joshi, Gauri and Poor, H. Vincent},
  title     = {Tackling the Objective Inconsistency Problem in Heterogeneous Federated Optimization},
  booktitle = {Advances in Neural Information Processing Systems},
  volume    = {33},
  pages     = {7611--7623},
  year      = {2020}
}

@inproceedings{Reddi2021FedOpt,
  author    = {Reddi, Sashank J. and Charles, Zachary and Zaheer, Manzil and Garrett, Zachary and Rush, Keith and Kone{\v{c}}n{\'y}, Jakub and Kumar, Sanjiv and McMahan, H. Brendan},
  title     = {Adaptive Federated Optimization},
  booktitle = {International Conference on Learning Representations},
  year      = {2021}
}

@inproceedings{Blanchard2017,
  author    = {Blanchard, Peva and El Mhamdi, El Mahdi and Guerraoui, Rachid and Stainer, Julien},
  title     = {Machine Learning with Adversaries: Byzantine Tolerant Gradient Descent},
  booktitle = {Advances in Neural Information Processing Systems},
  volume    = {30},
  year      = {2017}
}

@inproceedings{Yin2018,
  author    = {Yin, Dong and Chen, Yudong and Kannan, Ramchandran and Bartlett, Peter},
  title     = {Byzantine-Robust Distributed Learning: Towards Optimal Statistical Rates},
  booktitle = {Proceedings of the 35th International Conference on Machine Learning},
  series    = {Proceedings of Machine Learning Research},
  volume    = {80},
  pages     = {5650--5659},
  year      = {2018},
  publisher = {PMLR}
}

@inproceedings{ElMhamdi2018,
  author    = {El Mhamdi, El Mahdi and Guerraoui, Rachid and Rouault, S{\'e}bastien},
  title     = {The Hidden Vulnerability of Distributed Learning in Byzantium},
  booktitle = {Proceedings of the 35th International Conference on Machine Learning},
  series    = {Proceedings of Machine Learning Research},
  volume    = {80},
  pages     = {3521--3530},
  year      = {2018},
  publisher = {PMLR}
}

@article{Pillutla2022,
  author  = {Pillutla, Krishna and Kakade, Sham M. and Harchaoui, Zaid},
  title   = {Robust Aggregation for Federated Learning},
  journal = {IEEE Transactions on Signal Processing},
  volume  = {70},
  pages   = {1142--1154},
  year    = {2022},
  doi     = {10.1109/TSP.2022.3153135}
}

@inproceedings{Cao2021,
  author    = {Cao, Xiaoyu and Fang, Minghong and Liu, Jia and Gong, Neil Zhenqiang},
  title     = {{FLTrust}: Byzantine-Robust Federated Learning via Trust Bootstrapping},
  booktitle = {Network and Distributed System Security Symposium},
  year      = {2021}
}

@inproceedings{Karimireddy2022Bucketing,
  author    = {Karimireddy, Sai Praneeth and He, Lie and Jaggi, Martin},
  title     = {Byzantine-Robust Learning on Heterogeneous Datasets via Bucketing},
  booktitle = {International Conference on Learning Representations},
  year      = {2022}
}

@inproceedings{Bagdasaryan2020,
  author    = {Bagdasaryan, Eugene and Veit, Andreas and Hua, Yiqing and Estrin, Deborah and Shmatikov, Vitaly},
  title     = {How To Backdoor Federated Learning},
  booktitle = {Proceedings of the Twenty Third International Conference on Artificial Intelligence and Statistics},
  series    = {Proceedings of Machine Learning Research},
  volume    = {108},
  pages     = {2938--2948},
  year      = {2020},
  publisher = {PMLR}
}

@inproceedings{Guo2017,
  author    = {Guo, Chuan and Pleiss, Geoff and Sun, Yu and Weinberger, Kilian Q.},
  title     = {On Calibration of Modern Neural Networks},
  booktitle = {Proceedings of the 34th International Conference on Machine Learning},
  series    = {Proceedings of Machine Learning Research},
  volume    = {70},
  pages     = {1321--1330},
  year      = {2017},
  publisher = {PMLR}
}

@inproceedings{Li2020Fairness,
  author    = {Li, Tian and Sanjabi, Maziar and Beirami, Ahmad and Smith, Virginia},
  title     = {Fair Resource Allocation in Federated Learning},
  booktitle = {International Conference on Learning Representations},
  year      = {2020}
}

\section*{Author Biographies}

\renewcommand{\arraystretch}{1.2}
\noindent\begin{tabular}{@{}p{0.17\textwidth} p{0.78\textwidth}@{}}

\begin{minipage}[t]{\linewidth}
\vspace{0pt}
\includegraphics[width=\linewidth]{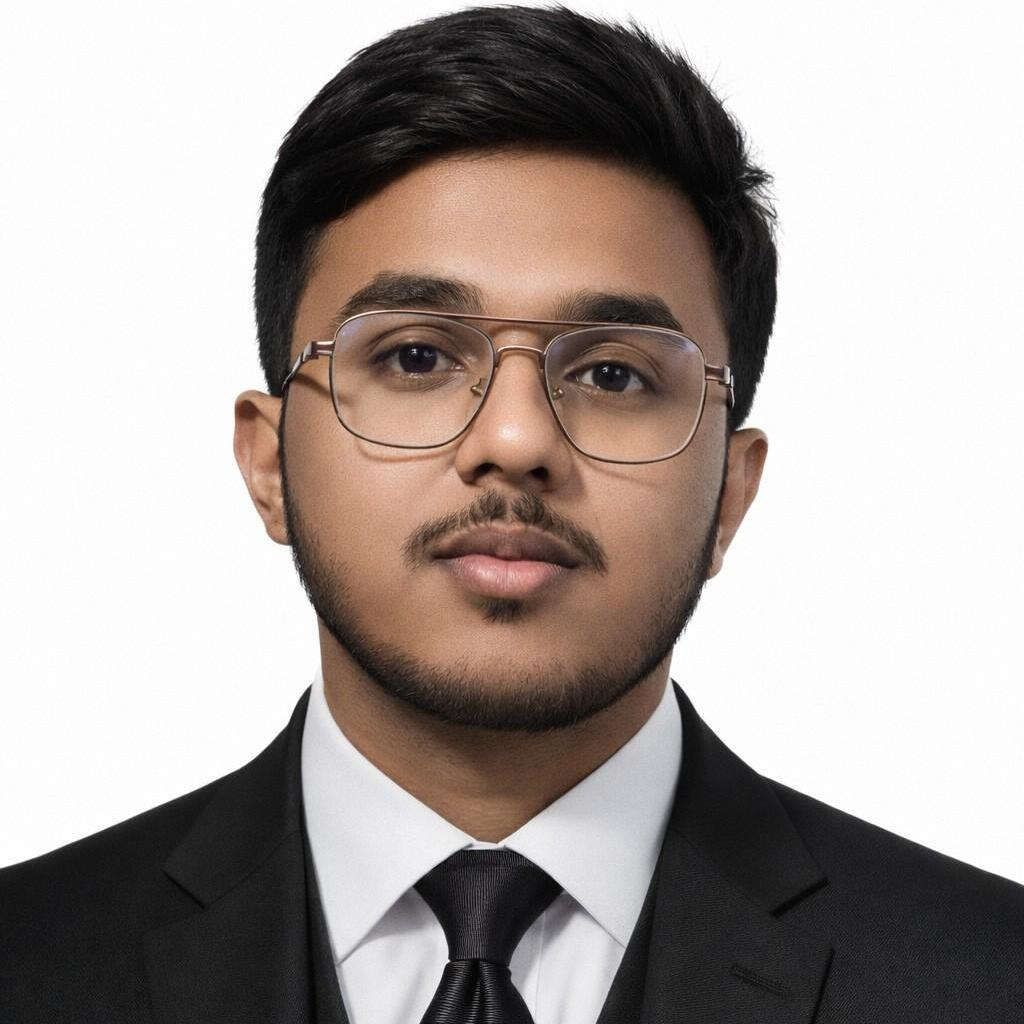}
\end{minipage}
&
\begin{minipage}[t]{\linewidth}
\vspace{0pt}
\textbf{Soumya Mazumdar} is a student researcher pursuing a B.S. in Data Science and Applications at the Indian Institute of Technology Madras and B.Tech. in Computer Science and Business Systems at West Bengal University of Technology (GMIT campus), India. His work focuses on temporal generative modeling, geometry-aware computer vision, and controllable video synthesis, with particular interest in diffusion-based methods for talking-head generation and temporal consistency. He has served as a Research Trainee at the Variable Energy Cyclotron Centre (VECC), where he worked on pose- and landmark-conditioned video generation, benchmarking, and efficient deployment pipelines. He has contributed to research publications in journals, conference proceedings, and edited volumes, and is also associated with an Indian patent in neural network-based real-time analysis.
\end{minipage}
\\[1.5em]

\begin{minipage}[t]{\linewidth}
\vspace{0pt}
\includegraphics[width=\linewidth]{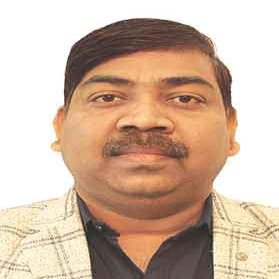}
\end{minipage}
&
\begin{minipage}[t]{\linewidth}
\vspace{0pt}
\textbf{Vineet Kumar Rakesh} is a Technical Officer (Scientific Category) at the Variable Energy Cyclotron Centre (VECC), Department of Atomic Energy, India, with over 23 years of experience in software engineering, database systems, and artificial intelligence. His research focuses on talking head generation, lip reading, and ultra-low-bitrate video compression for real-time teleconferencing. He is currently pursuing a Ph.D. at Homi Bhabha National Institute, Mumbai. Mr. Rakesh has contributed to office automation, OCR systems, and digital transformation projects at VECC. He is an Associate Member of the Institution of Engineers (India) and a recipient of the DAE Group Achievement Award.
\end{minipage}
\\[1.5em]

\begin{minipage}[t]{\linewidth}
\vspace{0pt}
\includegraphics[width=\linewidth]{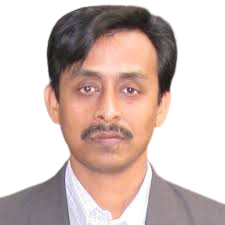}
\end{minipage}
&
\begin{minipage}[t]{\linewidth}
\vspace{0pt}
\textbf{Dr. Tapas Samanta} is a senior scientist and Head of the Computer and Informatics Group at the Variable Energy Cyclotron Centre (VECC), Department of Atomic Energy, India. With over two decades of experience, his work spans artificial intelligence, industrial automation, embedded systems, high-performance computing, and accelerator control systems. He also leads technology transfer initiatives and public scientific outreach at VECC.
\end{minipage}
\\
\end{tabular}

\vspace*{\fill}
\clearpage

\includepdf[
    pages=-,
    pagecommand={}
]{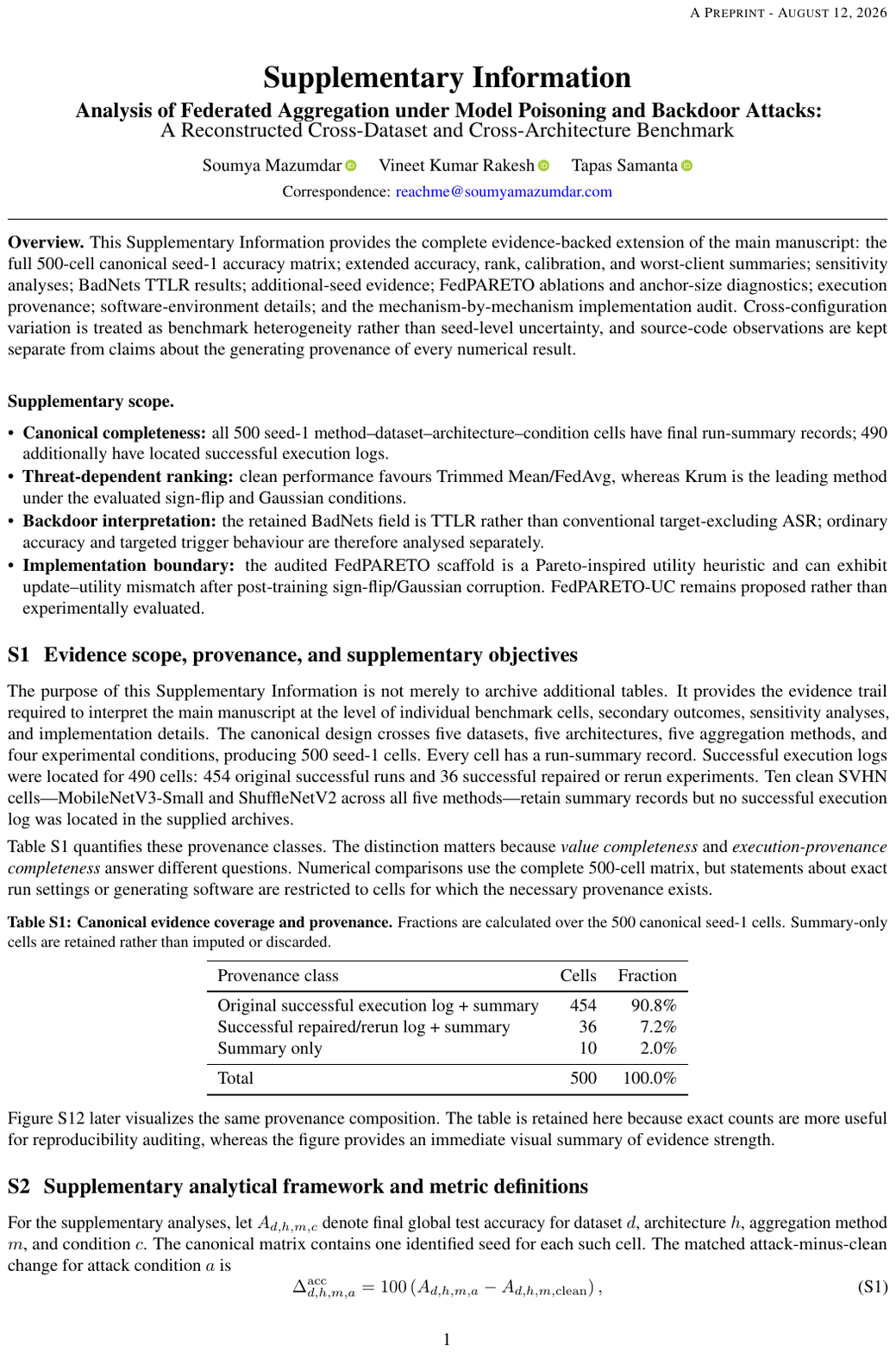}

\end{document}